\documentstyle[fleqn]{article}

\title{Integrating a Lexical Database and a Training Collection for
Text Categorization\thanks{This research is supported by the Spanish
Commitee of Science and Technology (CICYT TIC94-0187) - Published in
the {\em ACL/EACL Workshop on Automatic Extraction and Building of
Lexical Semantic Resources for Natural Language Applications, 1997}}}

\author{Jos\'e Mar\'{\i}a G\'omez-Hidalgo \and Manuel de
Buenaga Rodr\'{\i}guez\\
\{jmgomez,mbuenaga\}@dia.ucm.es\\
Departamento de Inform\'atica y Autom\'atica\footnote{Now Departamento
de Sistemas Inform\'aticos y Programaci\'on}\\
Universidad Complutense de Madrid\\
Avda. Complutense s/n, 28040 Madrid (Spain)}
\date{}
\begin{document}

\maketitle

\begin{abstract}
Automatic text categorization is a complex and useful task for many
natural language processing applications.  Recent approaches to text
categorization focus more on algorithms than on resources involved in
this operation.  In contrast to this trend, we present an approach
based on the integration of widely available resources as lexical
databases and training collections to overcome current limitations of
the task.  Our approach makes use of {\sc WordNet} synonymy
information to increase evidence for bad trained categories.  When
testing a direct categorization, a {\sc WordNet} based one, a training
algorithm, and our integrated approach, the latter exhibits a better
perfomance than any of the others.  Incidentally, {\sc WordNet} based
approach perfomance is comparable with the training approach one.
\end{abstract}

\section{Introduction}

{\em Text categorization} (TC) is the classification of documents with
respect to a set of one or more pre-existing categories.  TC is a hard
and very useful operation frequently applied to the assignment of
subject categories to documents, to route and filter texts, or as a
part of natural language processing systems.

In this paper we present an automatic TC approach based on the use of
several linguistic resources.  Nowadays, many resources like {\em
training collections} and {\em lexical databases} have been
successfully employed for text classification tasks \cite{boguraev96},
but always in an isolated way.  The current trend in the TC field is
to pay more attention to algorithms than to resources.  We believe
that the key idea for the improvement of text categorization is
increasing the amount of information a system makes use of, through
the integration of several resources.

We have chosen the Information Retrieval vector space model for our
approach.  Term weight vectors are computed for documents and
categories employing the lexical database {\sc WordNet} and the
training subset of the test collection Reuters-22173.  We calculate
the weight vectors for:

\begin{itemize}

\item A direct approach,

\item a {\sc Wordnet} based approach,

\item a training collection approach,

\item and finally, a technique for integrating {\sc WordNet} and a
training collection.

\end{itemize}

Later, we compare document-category similarity by means of a
cosine-based function.  We have driven a series of experiments on the
test subset of Reuters-22173, which yields two conclusions.  First, the
integrated approach performs better than any of the other ones,
confirming the hypothesis that the more informed a text classification
system is, the better it performs.  Secondly, the lexical database
oriented technique can rival with the training approach, avoiding the
necessity of cost-expensive building of training collections for any
domain and classification task.

\section{Task Description}

Given a set of documents and a set of categories, the goal of a
categorization system is to decide whether any document belongs to any
category or not.  The system makes use of the information contained in
a document to compute a degree of pertainance of the document to each
category.  Categories are usually subject labels like {\sc art} or
{\sc military}, but other categories like text genres are also
interesting \cite{karlgren94}.  Documents can be news stories, e-mail
messages, reports, and so forth.

The most widely used resource for TC is the training collection.  A
{\em training collection} is a set of manually classified documents
that allows the system to guess clues on how to classify new unseen
documents.  There are currently several TC test collections, from
which a training subset and a test subset can be obtained.  For
instance, the huge TREC collection \cite{harman96}, OHSUMED
\cite{hersh94} and Reuters-22173 \cite{lewis92} have been collected
for this task.  We have selected Reuters because it has been used in
other work, facilitating the comparison of results.

Lexical databases have been rarely employed in TC, but several
approaches have demonstrated their usefulness for term classification
operations like {\em word sense disambiguation}
\cite{resnik95,agirre96}.  A {\em lexical database} is a reference
system that accumulates information on the lexical items of one o
several languages.  In this view, machine readable dictionaries can
also be regarded as primitive lexical databases.  Current lexical
databases include {\sc WordNet} \cite{miller95}, EDR \cite{yokoi95}
and Roget's Thesaurus.  {\sc WordNet}'s large coverage and frequent
utilization has led us to use it for our experiments.

We organize our work depending on the kind and number of resources
involved.  First, a direct approach in which only the categories
themselves are the terms used in representation has been tested.
Secondly, {\sc WordNet} by itself has been used for increasing the
number of terms and so, the amount of predicting information.
Thirdly, we have made use of the training subset of Reuters to obtain
the categories representatives.  Finally, we have employed both {\sc
WordNet} and Reuters to get a better representation of undertrained
categories.

\section{Integrating Resources in the Vector Space \\ Model}

The {\em Vector Space Model} (VSM) \cite{salton83} is a very suitable
environment for expressing our approaches to TC: it is supported by
many experiences in text retrieval \cite{lewis92,salton89}; it allows
the seamless integration of multiple knowledge sources for text
classification; and it makes it easy to identify the role of every
knowledge source involved in the classification operation.  In the
next sections we present a straightforward adaptation of the VSM for
TC, and the way we use the chosen resources for calculating several
model elements.

\subsection{Vector Space Model for Text Categorization}

The bulk of the VSM for Information Retrieval (IR) is representing
natural language expressions as term weight vectors. Each weight
measures the importance of a term in a natural language expression,
which can be a document or a query. Semantic closeness between
documents and queries is computed by the cosine of the angle between
document and query vectors.

Exploiting an obvious analogy between queries and categories, the
latters can be represented by term weight vectors. Then, a category
can be assigned to a document when the cosine similarity between
them exceeds a certain threshold, or when the category is highly
ranked. In a closer look, and given three sets of $N$ terms, $M$
documents and $L$ categories, the weight vector for document $j$ is
$\langle wd_{1j}, wd_{2j}, \ldots , wd_{Nj} \rangle$ and the weight
vector for category $k$ is $\langle wc_{1k}, wc_{2k}, \ldots , wc_{Nk}
\rangle$. The similarity between document $j$ and category $k$ is
obtained with the formula:

\[sim(d_{j},c_{k}) = \frac{{\displaystyle \sum_{i=1}^{N} wd_{ij}
\cdot wc_{ik}}}{{\displaystyle \sqrt{{\sum_{i=1}^{N} wd_{ij}^{2}}
\cdot {\sum_{i=1}^{N} wc_{ik}^{2}}}}}\]

Term weights for document vectors can be computed making use of well
known formulae based on term frequency. We use the following one from
\cite{salton89}:

\[wd_{ij}=tf_{ij} \cdot \log_{2} \frac{M}{df_{i}}\]

Where $tf_{ij}$ is the frequency of term $i$ in document $j$, and
$df_{i}$ is the number of documents in which term $i$ occurs. Now,
only weights for category vectors are to be obtained. Next we will
show how to do it depending on the resource used.

\subsection{Direct Approach}

This approach to TC makes no use of any resource apart to the
documents to be classified.  It tests the intuition that the name of
content-based categories is a good predictor for the occurrence of
these categories.  For instance, the occurrence of the word ``barley''
in a document suggests that this one should be classified in the {\sc
barley}\footnote{All the following examples are taken from the Reuters
category set and involve words that actually occur in the documents.}
category.  We have taken exactly the categories names, although
classification in more general categories like {\sc strategic-metal}
should rather relay on the occurrence of more specific words like
``gold'' or ``zinc.''

In this approach, the terms used for the representation are just the
categories themselves.  The weight of term $i$ in the vector for
category $k$ is 1 if \(i=k\) and 0 in other cases.  Multiword
categories imply the use of multiword terms.  For example, the
expression ``balance of payments'' is considered as one term.  When
categories consist of several synonyms (like {\sc iron-steel}), all of
them are used in the representation.  Since the number of categories
in Reuters is 135, and two of them are composite, these approach
produces 137-component vectors.

\subsection{{\sc WordNet}-based Approach}

Lexical databases contain many kinds of information (concepts;
synonymy and other lexical relations; hyponymy and other conceptual
relations; etc.). For instance, {\sc WordNet} represents concepts as
synonyms sets, or {\em synsets}. We have selected this synonymy
information, performing a ``category expansion'' similar to query
expansion in IR. For any category, the synset it belongs to is
selected, and any other term belonging to it is added to the
representation. This technique increases the amount of evidence used
to predict category occurrence.

Unfortunately, the disambiguation of categories with respect to {\sc
WordNet} concepts is required. We have performed this task manually,
because the small number of categories in the test collection
made it affordable. We are currently designing algorithms for
automating this operation.

After locating categories in {\sc WordNet}, a term set containing all
the category's synonyms has been built.  For the 135 categories used
in this study, we have produced 368 terms.  Although some meaningless
terms occur and could be deleted, we have developed no automatic
criteria for this at the moment.

Let us take a look to one example.  The {\sc fuel} category has driven
us to the addition of the terms ``combustible'' and ``combustible
material,'' since they belong to the same synset in {\sc WordNet}.  In
general, the term weight vector for category $k$ is 1 for every
synonym of the category an 0 for any other term.

\subsection{Training Collection Approach}

The key asumption when using a training collection is that a term
often occurring within a category and rarely within others is a good
predictor for that category.  A set of predictors is typically
computed from term to category co-ocurrence statistics, as a training
step.  The computation depends on the approach and algorithm selected.
As Lewis has done before \cite{lewis92}, we have replicated in the VSM
early Bayesian experiments that had reported good results.

Terms are selected according to the number of times they occur within
categories. Those terms which co-occur at least with the 1\% and at
most with the 10\% of the categories are taken.  Among them, those 286
with higher document frequency are selected. We work the weights
out in the same way as in documents vectors:

\[wc_{ik}=tf_{ik} \cdot \log_{2} \frac{L}{cf_{i}}\]

Where $tf_{ik}$ is the number of times that term $i$ occurs within
documents assigned to category $k$, and $cf_{i}$ is the number of
categories within term $i$ occurs.  For example, after selecting and
weighting categories, the high-frequency term ``export'' shows its
largest weight for category {\sc trade}, but it also shows large
weights for {\sc grain} or {\sc wheat}, and small weights for {\sc
belgian-franc} and {\sc wool}.  A less frequent term typically
provides evidence for a smaller number of categories.  For example,
``private'' has a large weight only for {\sc acquisition}, and medium
for {\sc earnings} and {\sc trade}.

\subsection{Integrating {\sc WordNet} and a Training Collection}

Several ways of integrating {\sc WordNet} and Reuters have occurred to
us.  A sensible one is to use concepts instead of terms as
representatives.  However, and although promising, Voorhees reported
no improvements with this idea \cite{voorhees93}.  On the other side,
we have realized that the shortcomings in training can be corrected
using {\sc WordNet} to provide better forecast of low frequency
categories.

In general, we have linked {\sc WordNet} weight vectors to training
weigth vectors.  First we have removed those {\sc WordNet} terms not
ocurring in the training collection.  Then we have normalized both {\sc
WordNet} vectors and training vectors to separately add up across each
category.  This way we have smoothed training weights (much larger
than {\sc WordNet} ones), giving equal influence to each kind of term
weight.  This technique results in 461 term weights vectors, 185
coming from {\sc WordNet}, and 286 from training.  Weights for terms
ocurring in both sets have been summed.  Examples of terms coming from
training are ``import'' or ``government,'' with high weights for
highly frequent categories, like {\sc acq} (acquisition).  Examples of
terms coming from {\sc WordNet} are ``petroleum'' or ``peanut,'' with
weights only for the corresponding categories {\sc crude} and {\sc
groundnut} respectively.

We can clearly identify the role of each resource in this TC
approach. {\sc WordNet} supplies information on the semantic
relatedness of terms and categories when training data is no longer
available or reliable. It directly contributes with part of the terms
used in the vector representation. On the other side, the training
collection supplies terms for those categories that are better trained.
The problem of unavailability of training data is then overcome
through the use of an extern resource.

\section{Evaluation}

Evaluation of TC and other text classification operations exhibits
great heterogeneity.  Several metrics and test collections have been
used for different approaches or works.  This results in a lack of
comparability among the approaches, forcing to replicate experiments
from other researchers.  Trying to minimize this problem, we have
chosen a set of very extended metrics and a frequently used free test
collection for our work.  The metrics are {\em recall} and {\em
precision}, and the test collection is, as introduced before,
Reuters-22173.  Before stepping into the actual results, we provide a
closer look to these elements.

\subsection{Evaluation metrics}

The VSM promotes recall and precision based evaluation, but there are
several ways of calculating or even defining them.  We focus on
recall, being the discussion analogous for precision.  First,
definition can be given regarding categories or documents
\cite{larkey96}.  Second, computation can be done {\em macro-averaging}
or {\em micro-averaging} \cite{lewis92}.

\begin{itemize}

\item Recall can be defined as the number of correctly assigned
documents to a category over the number of documents to be correctly
assigned to the category.  But a document-oriented definition is also
possible: the number of correctly assigned categories to a document
over the number of correct categories to be assigned to the document.
This later definition is more coherent with the task, but the former
allows to identify the most problematic categories.

\item Macro-averaging consists of computing recall and precision for
every item (document or category) in one of both previous ways, and
averaging after it.  Micro-averaging is adding up all numbers of
correctly assigned items, items assigned, and items to be assigned,
and calculate only one value of recall and precision.  When
micro-averaging, no distinction about document or category orientation
can be made.  Macro-averaging assigns equal weight to every category,
while micro-averaging is influenced by most frequent categories.

\end{itemize}

Evaluation depends finally on the category assignement strategy:
probability thresholding, {\em k-per-doc} assignment, etc.  Strategies
define the way to produce recall/precision tables.  For instance, if
simmilarities are normalized to the $[0,1]$ interval, eleven levels of
probability threshold can be set to 0.0, 0.1, and so.  When the system
performs {\em k-per-doc} assignment, the value of $k$ is ranged from 1
to a reasonable maximum.

We must assign an unknown number of categories to each document in
Reuters.  So, the probability thresholding approach seems the most
sensible one.  We have then computed recall and precision for eleven
levels of threshold, both macro and micro-averaging.  When
macro-averaging, we have used the category-oriented definition of
recall and precision.  After that, we have calculated averages of
those eleven values in order to get single figures for comparison.

\subsection{The Test Collection}

The Reuters-22173 collection consists of 22,173 newswire articles from
Reuters collected during 1987.  Documents in Reuters deal with
financial topics, and were classified in several sets of financial
categories by personnel from Reuters Ltd.  and Carnegie Group Inc.
Documents vary in length and number of categories assigned, from 1
line to more than 50, and from none categories to more than 8.  There
are five sets of categories: TOPICS, ORGANIZATIONS, EXCHANGES, PLACES,
and PEOPLE. As others before, we have selected the 135 TOPICS for our
experiments.  An example of news article classified in {\sc bop}
(balance of payments) and {\sc trade} is shown in
Figure~\ref{reuters1}.  Some spurious formatting has been removed from
it.

\begin{figure}
\small
\begin{verbatim}
     PATTERN-ID 6505 TRAINING-SET
    18-JUN-1987 11:44:27.20
    TOPICS:     bop trade  END-TOPICS
    PLACES:     italy      END-PLACES
    PEOPLE:                END-PEOPLE
    ORGS:                  END-ORGS
    EXCHANGES:             END-EXCHANGES
    COMPANIES:             END-COMPANIES
    ITALIAN BALANCE OF PAYMENTS IN DEFICIT IN MAY
        ROME, June 18 - Italy's overall balance of payments showed
    a deficit of 3,211 billion lire in May compared with a surplus
    of 2,040 billion in April, provisional Bank of Italy figures
    how.
        The May deficit compares with a surplus of 1,555 billion
    lire in the corresponding month of 1986.
       For the first five months of 1987, the overall balance of
    payments showed a surplus of 299 billion lire against a deficit
    of 2,854 billion in the corresponding 1986 period.
     REUTER
\end{verbatim}
\caption{Document number 6505 from Reuters.} \label{reuters1}
\end{figure}

When a test collection is provided, it is customary to divide it into
a training subset and a test subset.  Several partitions have been
suggested for Reuters \cite{lewis92}, among which ones we have opted
for the most general and difficult one.  First 21,450 news stories are
used for training, and last 723 are kept for testing.  We summarize
significative differences between test and training sets in
Figure~\ref{reuters2}.  These differences can bring noise into
categorization, because training relies on similarity between
training and test documents.  Nevertheless, this 21,450/723 partition
has been used before \cite{lewis92,hayes90} and involves the general
case of documents with no categories assigned.

\begin{figure}
\begin{center}
\begin{tabular}{l|l|r|r|r} \hline\hline
\multicolumn{2}{c|}{}
  & \multicolumn{3}{c}{\em Subcollection} \\ \cline{3-5}
\multicolumn{2}{c|}{}
  & {\em Training} & {\em Test} & {\em Total} \\ \hline
Docs. & Number & 21,450 & 723 & 22,173 \\
Words & Occurrences & 2,851,455 & 140,922 & 2,992,377 \\
 & Doc. average & 127 & 195 & 134 \\
Docs. with 1+ Topics & Number & 11,098 & 566 & 11,664 \\
 & Percentage & 52 & 78 & 53 \\
Topics & Occurrences & 13,756 & 896 & 14,652 \\
 & Doc. Average & 0.64 & 1.24 & 0.66 \\
\hline\hline
\end{tabular}
\end{center}
\caption{\label{reuters2} Reuters-22173 document collection
statistics.}
\end{figure}

We have worked with raw data provided in the Reuters distribution.
Control characters, numbers and several separators like `/' have been
removed, and categories different from the TOPICS set have been
ignored.  For disambiguating categories with respect to {\sc WordNet}
senses, we first had to acquire their meaning, not always
self-evident.  This task has been performed by direct examination of
training documents.

\subsection{Results and Interpretation}

The results of our first series of experiments are summarized in the
table in Figure~\ref{results}. This table shows recall and
precision averages calculated both macro and micro-averaging for a
threshold-based assignment strategy. Values for the integrated
approach show some general advantage over {\sc WordNet} and training
approaches, but results are not decisive. Training results are
comparable with those from Lewis \cite{lewis92}, and the {\sc
WordNet} approach is roughly equivalent to the training one.

\begin{figure}
\begin{center}

\begin{tabular}{l|cc|cc} \hline\hline
{\em Threshold} & \multicolumn{2}{c|}{\em Macro-averaging} &
\multicolumn{2}{c}{\em Micro-averaging} \\ \cline{2-5}
{\em strategy} & {\em Recall} & {\em Precision} & {\em Recall} & {\em
Precision} \\ \hline
Direct & 0.239302 & 0.242661 & 0.205849 & 0.235775 \\
WordNet & 0.324899 & 0.306445  & 0.260762 & 0.298363 \\
Training & 0.325586 & 0.188701  & 0.365988 & 0.275731 \\
Integrated & 0.373365 & 0.220186 & 0.418652 & 0.296423  \\ \hline\hline
\end{tabular}
\end{center}
\caption{Overall results from our experiments.} \label{results}
\end{figure}

On one hand, the integrated approach shows a better performance than
the {\sc WordNet} one in general, although a problem of precision is
detected when macro-averaging.  The influence of low precison training
has produced this effect.  We are planning to strengthen {\sc WordNet}
influence to overcome this problem.  On the other hand, the integrated
approach reports better general perfomance than the training approach.

As expected, {\sc WordNet} and training both beat the direct approach.
When comparing {\sc WordNet} and training approaches, we observe that
the former produces better results with categories of low frequency,
while the latter perfoms better in highly frequent categories.
However, both exhibit the same overall behaviour.  Differences in
categories are noticed by the fact that micro-averaging is influenced
by highly frequent elements, while macro-averaging depends on the
results of many elements of low frequency.

\section{Related Work}

Text categorization has emerged as a very active field of research in
the recent years.  Many studies have been conducted to test the
accuracy of training methods, although much less work has been
developed in lexical database methods.  However, lexical databases and
especially {\sc WordNet} have been often used for other text
classification tasks, like word sense disambiguation.

Many different algorithms making use of a training collection have
been used for TC, including {\em k-nearest-neighbor} algorithms
\cite{masand92}, Bayesian classifiers \cite{lewis92}, learning
algorithms based in relevance feedback \cite{lewis96} or in decision
trees \cite{apte94}, or neural networks \cite{wiener95}. Apart from
\cite{lewis92}, the closest approach to ours is the one from Larkey
and Croft \cite{larkey96}, who combine {\em k-nearest-neighbor}, Bayesian
independent and relevance feedback classifiers, showing improvements
over the separated approaches. Although they do not make use of
several resources, their approach tends to increase the information
available to the system, in the spirit of our hypothesis.

To our knowledge, lexical databases have been used only once in TC.
Hearst \cite{hearst94} adapted a disambiguation algorithm by Yarowsky
using {\sc WordNet} to recognize category occurrences.  Categories
are made of {\sc WordNet} terms, which is not the general case of
standard or user-defined categories.  It is a hard task to adapt {\sc
WordNet} subsets to pre-existing categories, especially when they are
domain dependent.  Hearst's approach shows promising results confirmed
by the fact that our {\sc WordNet}-based approach performs at least
equally to a simple training approach.

Lexical databases have been employed recently in word sense
disambiguation.  For example, Agirre and Rigau \cite{agirre96} make
use of a semantic distance that takes into account structural factors
in {\sc WordNet} for achieving good results for this task.
Additionally, Resnik \cite{resnik95} combines the use of {\sc WordNet}
and a text collection for a definition of a distance for
disambiguating noun groupings.  Although the text collection is not a
training collection (in the sense of a collection of manually labelled
texts for a pre-defined text processing task), his approach can be
regarded as the most similar to ours in the disambiguation task.
Finally, Ng and Lee \cite{ng96} make use of several sources of
information inside a training collection (neighborhood, part of
speech, morfological form, etc.)  to get good results in
disambiguating unrestricted text.

We can see, then, that combining resources in TC is a new and
promising approach supported by previous research in this and other
text classification operations.  With more information extracted from
{\sc WordNet} and better training algorithms, automatic TC integrating
several resources could compete with manual indexing in quality, and
beat it in cost and efficiency.

\section{Conclusions and Future Work}

In this paper, we have presented a multiple resource approach for TC.
This approach integrates the use of a lexical database and a training
collection in a vector space model for TC. The technique is based on
improving the language of representation construction through the use
of the lexical database, which overcomes training deficiencies.  We
have tested our approach against training algorithms and lexical
database algorithms, reporting better results than both of these
techniques.  We have also acknowledged that a lexical database
algorithm can rival training algorithms in real world situations.

Two main work lines are open: first, we have to conduct new series of
experiments to check the lexical database and the combined approaches
with other more sophisticated training approaches; second, we will
extend the multiple resource technique to other text classification
tasks, like text routing or relevance feedback in text retrieval.


\begin{thebibliography}{boguraev96}

\bibitem{boguraev96}
Boguraev, B., Pustejovsky, J. (Eds.)
\newblock 1996.
\newblock {\em Corpus Processing for Lexical Acquisition}.
\newblock The MIT Press.

\bibitem{karlgren94}
Karlgren, J., Cutting, D.
\newblock 1994.
\newblock Recogninzing text genres with simple metrics using
discriminant analysis.
\newblock In {\em Proceedings of COLING'94}.

\bibitem{harman96}
Harman, D.
\newblock 1996.
\newblock Overview of the Forth Text Retrieval Conference (TREC-4).
\newblock In {\em Proceedings of the Fourth Text Retrieval Conference}.

\bibitem{hersh94}
Hersh, W, Buckley, C., Leone, T.J. Hickman, D. 1994.
\newblock 1994.
\newblock OHSUMED: an interactive retrieval evaluation and new large
test collection for research.
\newblock In {\em Proceedings of the ACM SIGIR'94}.

\bibitem{lewis92}
Lewis, D.D.
\newblock 1992.
\newblock {\em Representation and learning in information retrieval}.
\newblock Ph. D. Thesis, Dept. of Computer and Information Science,
University of Massachusetts.

\bibitem{resnik95}
Resnik, P.
\newblock 1995.
\newblock Disambiguating noun groupings with respect to WordNet
senses.
\newblock In {\em Proceedings of the Third Workshop on Very Large
Corpora}.

\bibitem{agirre96}
Agirre, E., Rigau, G.
\newblock 1996.
\newblock Word sense disambiguation using conceptual distance.
\newblock In {\em Proceedings of COLING'96}.

\bibitem{miller95}
Miller, G.
\newblock 1995.
\newblock WordNet: a lexical database for English.
\newblock {\em Communications of the ACM}, Vol. 38, No. 11.

\bibitem{yokoi95}
Yokoi, T.
\newblock 1995.
\newblock The EDR electronic dictionary.
\newblock {\em Communications of the ACM}, Vol. 38, No. 11.

\bibitem{salton83}
Salton, G., McGill, M.J.
\newblock 1983.
\newblock {\em Introduction to modern information retrieval}.
\newblock McGraw-Hill.

\bibitem{salton89}
Salton, G.
\newblock 1989.
\newblock {\em Automating text processing: the transformation,
analysis and retrieval of information by computer}.
\newblock Addison-Wesley.

\bibitem{voorhees93}
Voorhees, E.M.
\newblock 1993.
\newblock Using WordNet to disambiguate word senses for text retrieval.
\newblock In {\em Proceedings of the ACM SIGIR'93}.

\bibitem{larkey96}
Larkey, L.S., Croft, W.B.
\newblock 1996.
\newblock Combining classifiers in text categorization.
\newblock In {\em Proceedings of the ACM SIGIR'96}.

\bibitem{hayes90}
Hayes, P.J., Weinstein, S.P.
\newblock 1990.
\newblock {CONSTRUE/TIS:} a system for content-based indexing of a
database of news stories.
\newblock In {\em Proceedings of the Second Annual Conference on
Innovative Applications of Artificial Intelligence}.

\bibitem{masand92}
Masand, B., Linoff, G., Waltz, D.
\newblock 1992.
\newblock Classifying news stories using memory based reasoning.
\newblock In {\em Proceedings of the ACM SIGIR'92}.

\bibitem{lewis96}
Lewis, D.D., Schapire, R.E., Callan, J.P., Papka, R.
\newblock 1996.
\newblock Training algorithms for linear text classifiers.
\newblock In {\em Proceedings of the ACM SIGIR'96}.

\bibitem{apte94}
Apte, C., Damerau, F., Weiss, S.W.
\newblock 1994.
\newblock Automated learning of decision rules for text
categorization.
\newblock {\em ACM Transactions in Information Systems}, Vol. 12, No.
3.

\bibitem{wiener95}
Wiener, E.D., Pedersen, J., Weigend, A.S.
\newblock 1995.
\newblock A neural network approach to topic spotting.
\newblock In {\em Proceedings of the SDAIR'95}.

\bibitem{hearst94}
Hearst, M.
\newblock 1994.
\newblock {\em Context and structure in automated full-text
information access}.
\newblock Ph. D. Thesis, Computer Science Division, University of
California at Berkeley.

\bibitem{ng96}
Ng, H.T., Lee, H.B.
\newblock 1996.
\newblock Integrating multiple knowledge sources to disambiguate word
sense: an exemplar based approach.
\newblock In {\em Proceedings of the ACL'96}.

\end{thebibliography}
\end{document}